Ideas from Developmental Robotics and Embodied AI on the Questions of Ethics in Robots

Alexandre Pitti
ETIS laboratory, Université de Cergyy-Pontoise, UMR 8051 CNRS, ENSEA

I. INTRODUCTION

The development of Humanoid robots in Societies asks many questions in regard of their acceptance and their utility. Because of their realistic appearance but of their relatively poor capabilities to interact either physically or socially with persons, humanoid robots are at best rigid machines fully controllable and predictable but only in constrained environments, at worse fascinating automata that are mostly unpredictable or risky to manipulate due to their poor capacities of coordination and of adaptation in unconstrained environments. The usefulness of humanoid robots is still unclear at a short term but progresses in Artificial Intelligence appear an ineluctable movement in the long term to have more autonomous behaviors and more adaptivity. The question of ethics in robots is therefore legitimate as far as they will be in contact with persons. As any new technology, their introduction let no ones indifferent and carries many reactions (and fantasies) either over-pessimistic or over-optimistic about their real capabilities and potential risks. The types of interaction we may have can be ranged from unpredictable, unpleasant and frustrating interactions, to adaptive, unrepetitive and pleasant behaviors but how to know in advance if interacting wit these advanced toys can be a source of emancipation for us or a source of alienation? Everything cannot be anticipated but this may depend on person's mood, on the robot's use and purpose, but also on how they are designed. The development of Digital Technologies (Computers, Cell Phones and Internet) might be representative of how the robot business and its norms will evolve and be regulated in the future. Internet has already changed the way we communicate and the way we work.

It has changed also the manner we organize our social life and the delimitation of our sphere of privacy. As similar with other technologies, it is probable that a regulation at the society level will occur for robots in homes. For instance, as robots come to our houses, they can collect more detailed information about our habits than any other technologies can do, nonetheless the question of privacy is not yet asked and answered. And as different from other technologies, robots have the possibility to act almost autonomously in the physical and digital environment. As their body may resemble too much those of humans and their actions may appear intentional, they may generate also some troubles or injuries to persons in contact with, especially to persons not accustomed to like infants and the elderly people. Furthermore, as it is for video games, the sci-fi literature and the movies industry have popularized the idea that we will confuse ourselves and mix between reality and fiction concerning robots and that we will not be able to make any distinction with humans in the future. This is however not new and not specific to robots.

René Girard theorized it as the Mimesis or as the Mimetic Desire, this human trait by which we project to others our own emotions, desire and intentions (Girard, 1976, 1977). Girard had the insight that our autonomy is 'illusory' and our desires are always borrowed from the others. As Human Beings are always trapped into Mimesis, they are at the same time attracted to and repulsed by what resemble us and imitate us. In this state of mutual mimicry, it is difficult to understand who is the initiator and who is the imitator as the distance between the two vanishes. In the light of the Mimesis theory, humanoid robots and AI are our modern mirrors in which we project our fantasies.

We would like however to see in robots a more positive trend. As AI is a model of our own intelligence, these algorithms can serve at the end to understand us better, to gauge our limits and to help us to elevate our potentialities either physical or cognitive. They can be an opportunity for improving our learning capabilities and for assisting us in our daily life. As if we were in front of a new type of mirror, robots are reflective machines that ask us how we would define intelligence for ourselves or how we would like to see others to interact with us; it might serve to interrogate our own ethics. As robots can learn and improve their behaviors through interaction, they may at the end resemble us more, mimic our behaviors and learn our habits and our social rules; our bias and taboos.

More than any other machines, they may enhance our learning of things as we possess powerful implicit learning mechanisms that are affected by social interaction (Marshal & Meltzoff, 2015). We will develop in the following the question of the robot ethics in interaction with persons and to what extent ethical behaviors can be either coded or learned within one autonomous system. Robot Ethics is now an interdisciplinary research at the intersection of applied ethics and robotics. We would like to extend it to what is known in regard to human brain learning and infant development.

**Embodiment** – The way engineers conceive Intelligence as a direct consequence on how they design 'intelligent' behaviors in robots. This has at the end some impact on the robots' actions and interactions and therefore on its ethics. We think that Intelligence is not about the accumulation of data as a metaphor of a Storehouse or of the Computer but its about perceiving and deciding what that data means. And meaningful data cannot be dissociated from the physical support of this information: following Shannon theory of information, information has a physical reality. Since any knowledge is acquired only from the perceptual experiences we have within the environment, one key idea we would like to advance is that the Body is the support vector for developing meaningful information and Intelligence. The other is that information is obviously limited, incomplete, but also subjective. We think therefore that Embodied Intelligence (or Embodiment) is a prerequisite design principle for any agent to acquire any intelligent behaviors like ethical behaviors; see (Pfeifer & Scheier, 1999; Pfeifer & Bongard, 2006; Pfeifer & Pitti, 2012).

This proposal might be obvious or simplistic but it is only through the body that one agent can experience oneself, and others. One consequence is that we can see the Body as the medium by which a sense of Active Perception can be acquired. Active Perception is the ability to learn to predict and anticipate the future sensory outcomes of one's own actions and possibly those of others (Clark, 2015). This is at least what it is seen in infant cognitive development. Studies by Andrew Meltzoff and Jacqueline Nadel discovered how imitation is deeply rooten in our genes as infants learn culturally by imitating others starting even at birth; see (Nadel & Butterworth, 1999) and (Meltzoff & Decety, 2003). Andrew Meltzoff extended this idea and developed one paradigm known as the "like me" hypothesis saying that infants understand others actions as if they were "like them" through their own repertoire of actions acquired by embodiment (Meltzoff, 2007). By doing so, they can mentalize others intention through observation of their actions in the light of their owns and even at an early age. This proposal is an interesting working hypothesis to develop in robots to endow them with cognitive capabilities in order to understand humans intention through their bodies as if they were "like them". As one acquires more experiences about oneself and others, it can acquire a sense of what is Self and what is Others, which is one way to apprehend the question of Ethics.

We will see in other sections how Embodied Intelligence is linked to the way information is acquired and represented in the brain regions related to Agency, Self-Other representations, Empathy, Theory of Minds with the so-called Mirror Neurons System and how it has an impact on the way roboticists may design cognitive architectures in robots for ethical behaviors.

This digression on Embodiment was important to explain our approach and to emphasize the difference between what is called the weak AI, the idea that intelligent behaviors generated in a robot are simulated and handcrafted by engineers and strong AI, the idea that some aspects of intelligent behaviors can be learned and tested through the robot's own perceptual experiences. We believe that engineers should not code directly any abstract rules (e.g. ethical rules) within the robot's memory system like we would do in a computer's program and for old-fashion AI algorithms (including feed-forward algorithms like deep networks), because any situation is experienced through the bodily senses and should be learned in consequence through its specific sensorimotor connections. Because brain and body dynamics are so intricate, the embodied brain is open-ended, always in interaction within the environment whereas the computer is a closed system, working in abstract symbolic worlds outside of the physical reality (Lungarella & Sporns, 2005). Therefore the two strategies for computing information are different.

The way information is processed by the brain is distributed, asynchronous, analog and robust to noise; the complete opposite is true for computers in which information processing is centralized, synchronous, digital, symbolic and very sensitive to errors. Since the body serves as the interface from which any perceptual inference can be made, any robot anchors its perceptual memories transcribed into the 'language' spoken by its sensorimotor circuits. We believe that any formulation of ethical behaviors in robots should be written at the sensorimotor level through embodiment and not at a symbolic level.

**Development & Learning** – This difference between closed systems and open systems is important as far as we design active robots and as far as humans are in the loop. In line with Cybernetics (Wiener, 1948), Edgar Morin explains that what characterizes Complex Systems is their capabilities to organize their own behaviors through their actions, and by doing so, their actions modify their internal organizations (Morin & Le Moigne, 1999). This reflexivity of open-ended processes based on feedback is what makes any system adaptive and robust to variations. If any action for a robot may always introduce noise and randomicity, it gives rise also to new phenomenons proper to dynamical systems such as Emergence and Self-Organization important for adaptiveness and creativity.

If inter-acting is open-ended and not close, all situations are potentially novel. It is a co-constructive process as it changes oneself and others during time. This requires therefore some adaptive learning system to learn to anticipate and understand others' intention. Following this, according to Morin, any knowledge should conceive some subjectivity related to the observation of others to apprehend any ethical problem:

*"Toute connaissance (et conscience) qui ne peut concevoir l'individualité, la subjectivité, qui ne peut inclure l'observateur dans son observation, est infirme pour penser tous problèmes, surtout les problèmes éthiques. Elle peut être efficace pour la domination des objets matériels, le contraire des énèrgies et les manipulations sur le vivant. Mais elle est devenue myope pour appréhender les réalités humaines et elle devient une menace pour l'avenir humain."* Morin, E., Ethique (La méthode 6), Seuil, 2004, p. 65.

We think that this kind of reflexivity based on feedback is to what any autonomous learning systems have to tend to in order to transcribe information into a knowledge. As intrinsic noise and error are constitutive to any interaction with the environment even for robots, error-driven adaptation –or what is called reinforcement learning– may lead to more robust behaviors and to the detection of novel or unpredictable situations. Thinking about how agents evolve in a dynamical environment can help to design autonomous systems that can deal with uncertainties and with the unexpected. Hence, open-ended agents should display properties of learning by trial and errors in order to test situations from their actions. Following Alan Turing's proposal, any robot should learn incrementally just like a child does by interacting with humans through trial and errors and by imitation (Turing, 1950). In line with him, we propose to not code any rules into robots but make them to experience and learn ethical rules through interactions with persons as a developmental process. That is, for us ethics should be a result of the robot's own development and interactions with others and not something coded abstractly, separately programmed from the rest of its sensorimotor decision making system as it would be for one pluggable module. Developmental learning should lead therefore to adaptive behaviors for which ethics is one aspect, embodied and intertwined with its own actions and comprehension of the scene.

*"Instead of trying to produce a programme to simulate the adult mind, why not rather try to produce one which simulates the child's? If this were then subjected to an appropriate course of education, one would obtain the adult brain [...] Our hope is that there is so little mechanism in the child brain that something like it can be easily programmed. The amount of work in the education we can assume, as a first approximation, to be much the same as for the human child."* (Turing, 1950, pp.456)

This idea is at the ground of Developmental Robotics whose main goal is to model the development of increasingly complex cognitive processes in natural and artificial systems and to understand how such processes emerge through physical and social interaction (Asada & al. 2001; Metta & al., 2003). This approach sees the whole body as a unified computational machine in order to learn from actions any perceptual experiences including interacting with others. This objective relies on the discovery of the Mirror Neurons System within the brain (Rizzolatti & al., 2001), which puts a strong role on multimodal integration and sensorimotor integration for Agency and to the development of a Theory of Mind in infants (Gallese, 2005). Researchers in Developmental Robotics would like to replicate these cognitive architectures in robots to benefit from these possibilities (Kuniyoshi, 2014; Cangelosi & Schesinger, 2015). For instance, the roboticist Minoru Asada proposed a developmental scenario how ethical values like empathy can be shared and learned within a robot by developing a sense of Self linked to Others (Asada, 2015). At the root of any of these cognitive architectures, still speculative, is the capability to predict the consequences of oneself actions (Asada et al., 2011).

**Active Inference** – Being embodied implies to rely on timing of events, sensorimotor coordination, on perceiving that our own actions have some impacts in the world. Since every situation is different and depends on a context, learning in which state the robot is and adapting its plan in consequence, with respect to the current context or goal, is an important feature to endow any autonomous system. Similar to infants, this ability of 'self-calibration' as expressed by the researcher in cognitive development Philippe Rochat is crucial for realizing intelligent behaviors in robots and to develop in them a sense of control (Rochat, 2003, 2011). This sense of control on things or of agency gives in return more predictability to persons in interaction with and a sense of comfort as the machine behaves as expected. Agency or Self-judgment is based on error prediction signals that assess current sensory activity based on the brain's expectations. According to it, the brain is continuously attempting to minimize the discrepancy or prediction error in order to adapt itself to the current situation, and this corresponds to the feeling of Agency or Awareness.

This capability of the brain for coding prediction (Predictive Coding) is however important for scene comprehension, reading one mind, as well as for Self representation and also for interacting with others. Recently, many researchers are defending this idea, like Andy Clark (Clark, 2015), Karl Friston and others (Friston & Kiebel, 2009; Apps & Tsakiris, 2014). In so far, no robot can even roughly recognize a scene, itself on a mirror, grasp satisfyingly an object, understanding where its own hand is or learning to predict others' intention.

**Self-Other Brain Regions, the Mirror Neurons** – According to Anil Seth, the Self network in the human brain may learn interoceptive signals by infering hidden causes (active inference) as well as errors (error-learning) when they can(not) be predicted (Seth, 2013). Seth identified the Anterior Insular Cortex (AIC) as this comparator circuit to be engaged in interoceptive inference, which is useful for error learning for this Self Network. By extension, it is proposed that the comparator circuit in AIC may exhibit the development of a network for a Theory of Mind (mind-reading capability) using also predictive coding and error-learning but for exteroceptive inference (Frith & Frith, 2003). Through embodied interactions, we suggest that what we might think as high-level abstract rules (e.g., ethical rules) can be learned and created at a very low-level by imitating robots endowed with such cognitive architecture and based on predictive coding.

Other candidates brain regions have been investigated by neuroscientists for mind-reading capabilities, like the so-called Mirror Neurons system, which has been extensively modeled by developmental roboticists. The MNS is placed in the shared circuits in the monkey and the humans parietal, in the temporal and motor areas formed from reciprocal and anatomical connections that work in parallel for transforming sensorimotor information (Ferrari & al., 2009; Murata & Ishida, 2007). Although their primary functions are aimed at interacting in the physical world by processing the multimodal sensory information about objects into motor commands, modern neurosciences attribute them a far broader role to engage oneself into the social world.

The principal evidence comes from the finding by Rizzolatti and colleagues of a particular class of neurons in the monkey F5 motor area, which is firing both when the monkey executes one action and when he is observing someone else executing it (Rizzolatti, & al. 1996; Gallese et al., 1996). The metaphor of a mirroring mechanism between the sensory and motor apparatus for generating one action and for understanding those of others has been retained to name this special class of neurons found primarily in the motor circuits and then in the parietal cortex (Keysers, 2004; Gallese et al., 2004). These mirror neurons are multimodal neurons that merge signals from visual, proprioceptive, auditory and somatopic systems (Sakata, et al. 1997; Caggiano et al., 2009). Some of them (in the Ventral Intra-Parietal area) are involved in the representation of the space within reach, the peripersonal space, which encodes a body image at the skin surface aimed at locating the relative position of the body-parts and of the objects nearby in body-centered coordinates (Rizzolatti et al., 1997).

By extension, these mirror-like neurons describe not only how the body interacts physically in the environment but also how the bodily-self binds socially with others. These features of the mirror neurons (MN) have deep implications as they may furnish some grounds on how higher cognitive skills could have arisen from the neural extent of the body representation itself as it is argued for empathy (Decety & Sommerville, 2007; Decety & Jackson, 2004; Bufalari et al., 2007), theory of mind (Fogassi et al., 2005; Fujii et al., 2008), the roots of language (Rizzolatti & Arbib, 1998) and even corporeal awareness (Keysers & Gazzola, 2006; Rizzolatti & Fabbri-Destro, 2008).

There are therefore of some interests for modeling the features of these brain structures in robots, which may serve for supporting ethical behaviors.

**Explanatory Learning Algorithms** – Predictive Coding is an interesting paradigm for ethical robots because it has its roots with Bayesian Inference. Bayesian theory is used to update the probability for a hypothesis as more evidence or information becomes available. Under this framework, one robot may update its current hypothesis about the current states or context based on its incoming signals or based on its own actions in the environment. Any error or conflicts may update its self-judgment about this hypothesis, which may in return be explained further to one user. After all, it is only when we can explain something clearly to someone that we have a perfect understanding of it. This goes in the direction of designing robots to learn by doing and understanding through trial and errors, to discover causal rules as if they were little scientists, which is proposed that infants do (Gopnik et al., 2000, Meltzoff, 2007, Tenenbaum, 2011). Following Richard Feynman's quote *'I cannot understand what I cannot create'*, the principle of enaction, the organization of knowledge through action, is a necessary choice to have robots that can learn by doing.

In this sense, we can say that Bayesian-based algorithms are one instance of Explanatory Learning Algorithms. Since current hypothesis can be evaluated and expressed to the user, this type of learning is different from black-box learning algorithms in which the internals state is not accessible to users. Bayesian processes permit to qualify uncertainties by estimating quantitatively what is unexpected or unpredictable. Certain AI architectures are properly designed to infer hidden causes like Boltzmann machines, auto-encoders or the Bayesian Networks and new implementations have been proposed for reverse-engineering like the generative adversarial networks (Goodfellows et al., 2014), or some implementations of predictive coding (Spratling, 2016).

By inferring the hidden causes of the current state of the robot, predictive coding and the Bayesian framework can serve to design for tracing back robot decisions. This has a drawback, to let the robot to act and perhaps to fail in certain situations by testing different hypothesis. It is only at this price that predictive coding algorithms can promote open access to robots cognitive architecture, which can serve for a better understanding of the robot's internal state and a better control by a user.

**A Science of Learning** – In a seminal paper, some developmental scientists advocated for a "foundations of a new Science of Learning" that would be on the frontiers of eduction, medical, cognitive sciences and machine learning (Meltzoff & al., 2009).
As a mirror of ourselves, having robots that can learn incrementally from us some hypothesis about the world, under uncertainty and noise, through trial and errors, may serve us to figure out about our own errors and failures, as well as the biases we human unconsciously introduce and propagate.

This approach, more horizontal about the capabilities of robots and more open-ended about their interactions in their environment, might be less effective than current algorithms specialized in one specific task only. At reverse, they may generalize more easily and be more interesting to use in terms of novelty and co-construction with a partner. On the long run, we speculate that curiosity-driven algorithms for life-long learning in autonomous machines may have some beneficial effects also on people interacting with to improve their learning capabilities either for cognitive tasks or for sport practices, and not only to infants or to the ederly persons. We think that the complexity and adaptiveness of autonomous robots is at this price.

ACKNOWLEDGMENTS

Grants EQUIPEX-ROBOTEX (CNRS), chaire d'excellence CNRS-UCP and Labex MME-DII (ANR11-LBX-0023-01).